\documentclass[letterpaper, 10 pt, conference]{ieeeconf}  

\IEEEoverridecommandlockouts                              

\usepackage[table]{xcolor} 
\usepackage{caption} 

\usepackage{graphicx} 
\usepackage{amsmath} 
\usepackage{cite}
\usepackage{hyperref}
\usepackage{siunitx}

\newcommand{\mypara}[1]{\par\vspace*{1.5mm}\noindent\textbf{{#1}}}

\usepackage[square, numbers, sort&compress]{natbib}

\usepackage{booktabs}
\usepackage{tabularx}

\title{\LARGE \bf
Text2Robot: Evolutionary Robot Design from Text Descriptions
}

\author{Ryan P. Ringel$^{*}$, Zachary S. Charlick$^{*}$, Jiaxun Liu$^{*}$, Boxi Xia and Boyuan Chen
\thanks{This work is supported by DARPA FoundSci program under award HR00112490372, by ARL STRONG program under awards W911NF2320182 and W911NF2220113. All authors are from Duke University. $^{*}$Equal contribution.}}

\begin{document}

\maketitle
\thispagestyle{empty}
\pagestyle{empty}

\begin{abstract}

Robot design has traditionally been costly and labor-intensive. Despite advancements in automated processes, it remains challenging to navigate a vast design space while producing physically manufacturable robots. We introduce Text2Robot, a framework that converts user text specifications and performance preferences into physical quadrupedal robots. Within minutes, Text2Robot can use text-to-3D models to provide strong initializations of diverse morphologies. Within a day, our geometric processing algorithms and body-control co-optimization produce a walking robot by explicitly considering real-world electronics and manufacturability. Text2Robot enables rapid prototyping and opens new opportunities for robot design with generative models. Our website is at \url{http://generalroboticslab.com/Text2Robot/}.

\end{abstract}

\section{Introduction}

For over half a century, robot design has been a costly and labor-intensive process, requiring extensive human efforts from initial sketches to detailed modeling, prototyping, controller design, manufacturing, and testing. This traditional approach has significant limitations, such as prohibitive costs, lengthy development cycles, and constraints on innovation bounded by human imagination and manual capabilities. However, advancements in automated robot design \cite{Lipson2000,hornby2003generative, matthews2023efficient,wang2019neural} promise to revolutionize this landscape. By automating key aspects of the design process, we can drastically reduce development time and costs, allowing industries to rapidly produce specialized robots and enabling engineers to establish efficient manufacturing processes. Researchers also benefit by quickly innovating desired hardware platforms. Ultimately, automating robot design not only enables rapid prototyping but also expands the realm of possible innovations, surpassing the boundaries of what human designers can envision and create.

One major challenge in automating robot design is navigating the vast and intricate design space. Traditional engineering design is time-intensive and demands considerable technical expertise. While advancements in control engineering \cite{gehring2013control,carpentier2021recent, ding2021representation,bledt2018cheetah, Farshidian2017} and machine learning \cite{Wang2012, lee2020learning,   Tsounis2020, ibarz2021train, miki2022learning} have enabled automatic training of robot policies, designing morphologies remains laborious. Human designers typically spend months conceptualizing, designing, and fabricating a robot, balancing cost, manufacturability, and performance. Previous automation attempts often simplify the design space by using large repeating modules \cite{Lipson2000, strgar2024evolution, zhao2020robogrammar, Gupta2021} or voxel representations \cite{Cheney2014, cheney2015evolving}. Though innovative, these designs are slow due to the need to search within a vast design space and do not consider real-world fabrication, resulting in theoretically sound but impractical designs to produce.

\begin{figure}[t!]
\begin{center}
\includegraphics[width=0.9\columnwidth]{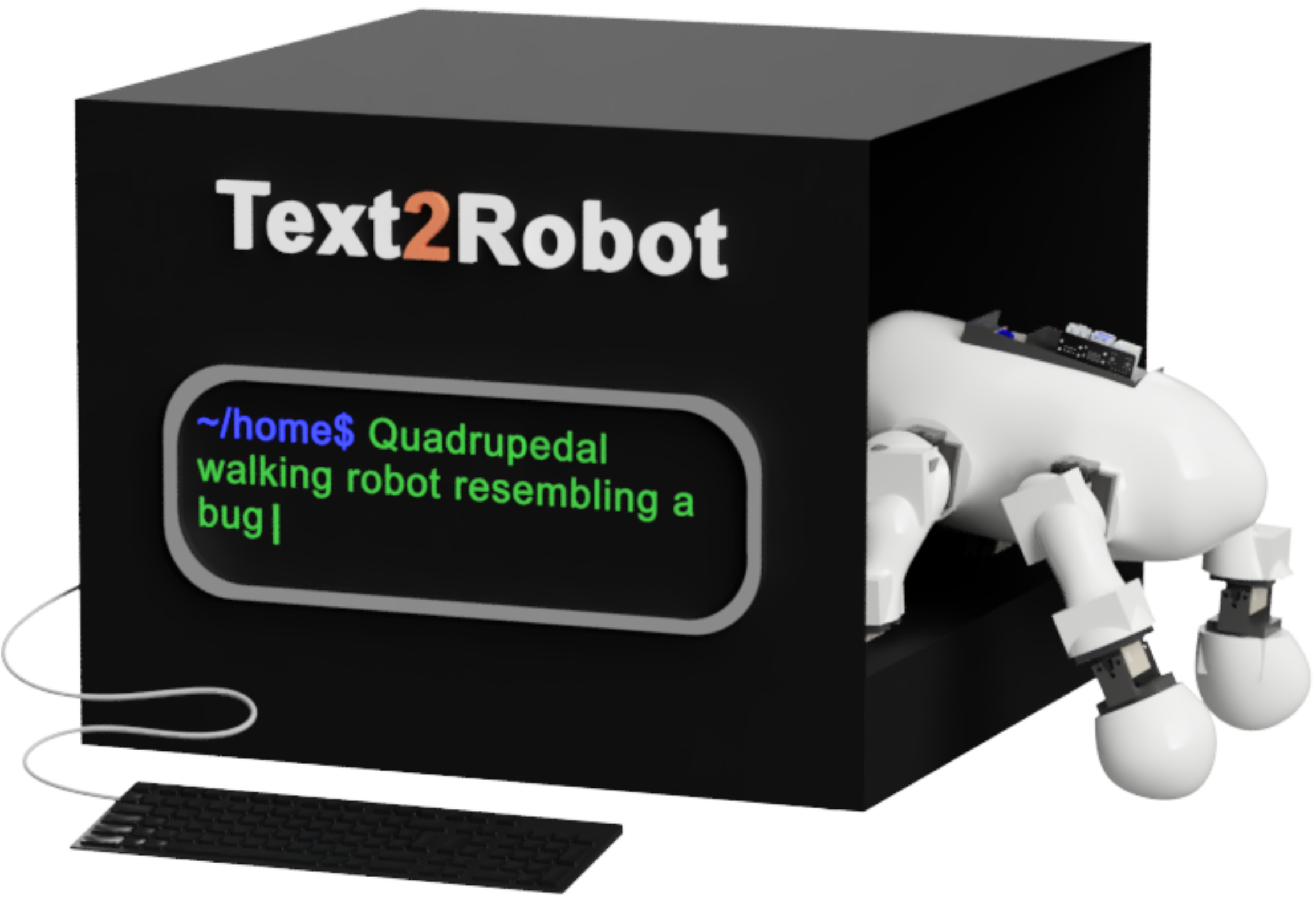}
\vspace{-3pt}
\caption{\textbf{Text2Robot} creates physical robots from user-specified text prompts and performance preferences while considering real-world electronics and manufacturability.}
\label{fig:enter-label}
\vspace{-28pt}
\end{center}  
\end{figure}

Automated methods for robot design predominately involve Evolutionary Algorithms (EAs) inspired by natural evolution \cite{back-eas}. However, EA-based approaches are inherently slow, starting with random solutions and iterating through hundreds of generations. Moreover, existing solutions do not scale well with increasing design complexity \cite{6792511} and face significant challenges in balancing multiple objectives, such as control and morphology co-optimization \cite{Gupta2021,Cheney2016}. Consequently, while these solutions may excel in simulations, they frequently fail to address practical issues such as sim2real transfer \cite{zhao2020robogrammar, Gupta2021, Cheney2014} and manufacturability. Problems like high current draw from unrealistic degrees of freedom \cite{zhao2020robogrammar, Gupta2021} or complex morphologies that are difficult to manufacture \cite{zhao2020robogrammar, Gupta2021, Cheney2014} hinder the transition from theoretical designs to practical and producible robots.

We present Text2Robot (Fig.~\ref{fig: pipeline}), an ``A-to-Z'' framework from user text specifications to physical walking robots. Our approach utilizes recent advancements in text-to-3D generative models to create initial mesh designs, which are subsequently converted into kinetic robot models through our geometric processing algorithms. Within minutes we can generate a design, within an hour, a robot trained in simulation, and within a day, a fabricated walking robot. Our system not only fulfills users' aesthetic preferences but further optimizes the designs using an evolutionary algorithm to incorporate other performance preferences. Our key insight is that text-to-3D generative models can provide a much stronger starting point for the evolutionary algorithm, significantly accelerating the optimization process. Experiments in both simulation and the physical world demonstrate our ability to specify both aesthetic qualities and performance metrics, such as velocity tracking and energy efficiency. Overall, our approach introduces the creative and artistic nature of generative models to automated robot design with fast prototyping from a text prompt, and has the potential to open up novel opportunities in rapid design and manufacturing.

\section{Related Work}

\mypara{Generative Models for Design} Generative AI aims to enhance design efficiency \cite{regenwetter2022deep,akande2024review} and reduce labor costs \cite{makatura2023can, feuerriegel-gen-ai} by automating tasks \cite{kazi2017dreamsketch} or portions of the workflows \cite{sanchez2018inverse, buonamici2020generative,Liu2023}. Recent advances in generative models have led to the integration of generative adversarial networks (GANs) \cite{liu2018generative,oh2019deep}, diffusion models \cite{luo2022antigen,yang2023diffusion}, and transformers \cite{wu2021deepcad, Siddiqui2023} into the generative design process. Previous research has utilized generative models for domain-specific code creation \cite{nordmann2014survey,chen2021evaluating} or parameter-based physical designs \cite{abdullah2013parametric, hornby2001advantages}. Such systems often heavily rely on human feedback and tuning during the design process. In contrast, our work focuses on directly enabling artificial intelligence-generated content (AIGC) for physically embodied and functional robot design.

\mypara{Text-to-3D Models} Recent advancements in text-to-3D generative models \cite{liu2024comprehensive} utilize pre-trained text-to-image diffusion models to optimize Neural Radiance Fields (NeRF) \cite{lin2023magic3d, poole2022dreamfusion,babu2023hyperfields}, text-to-3D shape embeddings in a GANs framework \cite{chen2019text2shape}, or explicit and hybrid scene representations \cite{chen2023fantasia3d} to create 3D generated content. Despite their growing capabilities, current text-to-3D models are primarily used for visualization in graphics and not for the creation of functioning machines. Our work incorporates 3D AIGC into physical robot design while considering electrical components and manufacturability constraints.

\mypara{Automated Robot Design} Automated robot design often involves co-optimizing control and morphology using classical methods and analytical dynamics \cite{Park1994, Paul2001, geijtenbeek2013flexible, ha2018computational}. These approaches rely on highly parameterized designs and specified models, which limit creativity and practicality in real-world environments. Recent studies employ reinforcement learning for co-optimization through optimizing the distribution of design parameters\cite{schaff2022soft,schaff2019jointly} or auto-differentiable hardware policies \cite{chen2020hardware, Xu-RSS-21, he202morph}, but their designs focus on limited design space and often overlook real-world manufacturability and electronics. Evolutionary Algorithms (EA) have been widely used \cite{Doncieux2015} to explore vast design spaces by combining modular building blocks \cite{Lipson2000, strgar2024evolution, zhao2020robogrammar, alattas2019evolutionary} or voxels \cite{Cheney2014, cheney2015evolving, Cheney2016} through genetic operators to co-evolve morphology and control to produce complex and unexpected \cite{Lehman2020} designs. However, EA-based approaches are computationally expensive, typically starting from random solutions and requiring numerous generations to produce effective designs. Our work leverages the open-ended nature of EA-based approaches for design evolution but drastically accelerates the process through strong initializations by incorporating generative models.

\section{Approach}

\begin{figure}[htbp]
\begin{center}
\vspace{5pt}
\includegraphics[width=0.95\columnwidth]{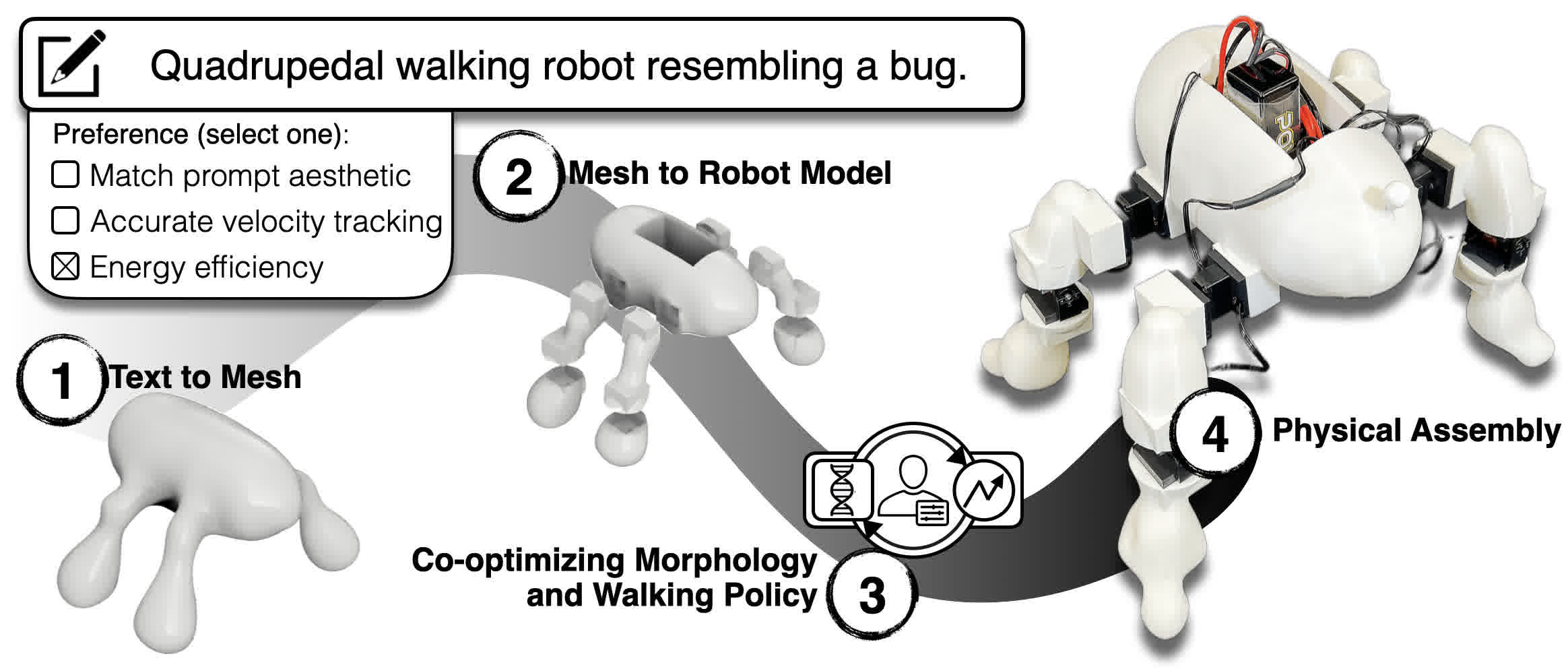}
\vspace{-5pt}
\caption{Overview of the four steps in Text2Robot framework.}
\label{fig: pipeline}
\vspace{-25pt}
\end{center}
\end{figure}

Text2Robot (Fig.~\ref{fig: pipeline}) generates a physical walking quadrupedal robot that caters to a user's text description and performance priorities, such as energy efficiency or velocity tracking accuracy. There are four major components in Text2Robot: (1) A generative model to create static 3D meshes of robots given user-specified texts. (2) A set of geometric processing algorithms to convert the static meshes into kinetic models, including the necessary components for fabrication. (3) An optimization process based on evolutionary algorithms and reinforcement learning to further optimize the robot morphologies and walking policies according to the user's performance preferences. (4) A final optimized robot is quickly 3D-printed and assembled.

\subsection{Mesh Generation from Text Prompts}

Given a text prompt specifying the aesthetic of the robot, we generate a 3D mesh of the robot using a text-to-3D model. In this paper, we used one of the state-of-the-art models, Meshy \cite{meshy}, that takes in the text prompt and produces several candidate meshes. One high-level assumption in this paper is to demonstrate our framework by automating the design of quadrupedal robots with eight motors. Having such constraints mimics the typical real-world design requirements without the loss of generality of our framework. We implemented a structured prompt design based on specified user descriptions to ensure Meshy consistently outputs quadrupedal meshes.

The user provides a text description of their desired robot in one to three words, which we incorporate into the following prompt format: \textless\texttt{Quadrupedal walking robot resembling a \textit{"User-Provided Description}"}\textgreater. The generated candidate meshes are then manually filtered according to the following constraints: (1) the mesh must be continuous without disjoint bodies; (2) the mesh must exhibit bilateral symmetry; and (3) the mesh must include four legs.

\subsection{Kinetic Robot Model from Static Meshes}

Current text-to-3D models only produce static meshes for visualization purposes. Our key challenge is to automatically convert such static meshes into kinetic robot models. Importantly, unlike most simulated robot models that are simplified for fast simulation, to automatically transfer our designs to physical functioning robots, our generated robot model should also consider real-world manufacturing factors such as the placement of electronic components, wire connections, physical collisions at joints, limits of the number of motors, and manufacturability.

\mypara{Mesh Repair and Preprocessing} Due to the lack of realistic constraints on the text-to-3D models, the generated mesh can have errors such as being non-watertight. We first call the mesh repair API through Fusion 360 \cite{Fusion360} to repair the mesh for the downstream workflow. We then leverage the mesh conversion operation to convert the mesh to an organic BREP (Boundary Representation) body. We scale the BREP body to a volume of $6300\si{\cm\cubed}$ to unify the initial mass of all robots.

\mypara{Joint Allocation} Deciding a set of feasible joint positions for a quadrupedal robot with eight motors purely from the mesh model is difficult. Inspired by natural quadrupedal animals, we assume that our robots have four legs and two movable joints for each leg. In other words, our robots have four shoulder joints and four knee joints, dividing each leg into an upper and lower leg.

\begin{figure}[t!]
\vspace{2pt}
\centering
\includegraphics[width=0.95\columnwidth]{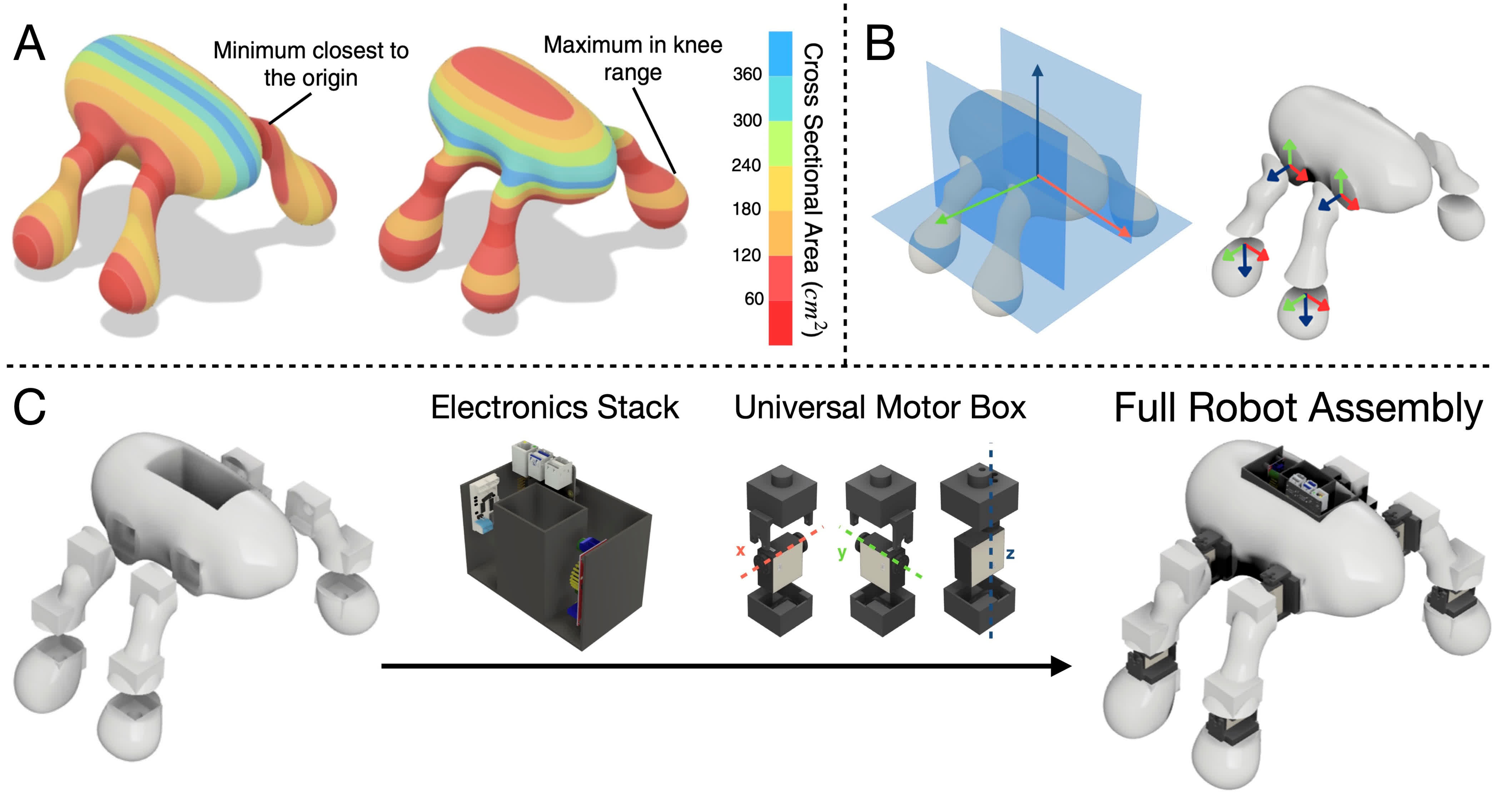}
\centering
\vspace{-3pt}
\caption{\textbf{Geometric Processing.} (A) Heat maps to visualize the cross-section area from the XZ (left) and XY plane (right). (B) The selected planes for slicing the mesh model and the coordinate of the center of mass and the resulting  robot components and their joint coordinates. (C) The final robot model with extruded boxes for electronics and motors.}
\vspace{-20pt}
\label{fig: geometric processing}
\end{figure}

We determine the position and orientation of each joint based on the mesh model's geometric features (Fig.~\ref{fig: geometric processing} A and B). The body's origin is defined as the center of mass, assuming uniform mass distribution. Starting from the origin along the $+y$ and $-y$ direction, we create vertical slices parallel to the $xz$ plane and record the cross-sectional area at each step. The slice closest to the origin with a local minimum in cross-sectional area is mirrored to the other side of the origin. The two slicing planes separate the mesh into four legs and one body, defined as the base link. The origins of the four shoulder joints correspond to the centroids of their intersection profiles, with $z$-axes perpendicular to the slicing plane and $y$-axes pointing downwards. Similarly, knee joints are located at the slice planes with maximum cross-sectional area, traversing along the $xy$ plane for each leg. We traverse from the bottom of the base link and stop $2\si{\cm}$ from the ground to ensure space for motor placement. The origins of the knee joints correspond to the centroids of each intersection profile, with $z$-axes perpendicular to their slicing planes and $x$-axes parallel to the slicing planes pointing in the robot's facing direction. We follow the right-hand system.

\mypara{Electronic Component Placement} To create robots for real-world manufacturing and walking, we need to consider the placements of electronic components, including actuators, batteries, and controllers. As in Fig.~\ref{fig: geometric processing}C, servo motors are housed in a 3D-printed box with snap-in pegs inspired by toy building blocks. Motors can be rotated in their casings to achieve the desired axis of rotation during assembly. A larger box encases other necessary electronics, such as the motor controller, Raspberry Pi, and battery, and can be easily slid into a central channel in the base link. To avoid self-collision and reserve space for motor and electronics modules, we offset the eight limbs by $4\si{\cm}$ and cut extrusions.

By this step, the robot's kinetic model is automatically defined by our algorithms, and our pipeline exports the model to Unified Robotics Description Format (URDF) \cite{toshinori2020fusion2urdf}. To enlarge the design space for optimization in Sec.~\ref{sec: optimization}, we scale each leg in $0.5\si{\cm}$ increments to generate nine additional models. We further augment these models to thirty variants by defining each joint's rotational axis along the $x, y, \textrm{or}~z$ axis of the joint coordinate.

\subsection{Co-Optimization of Morphology and Policy}
\label{sec: optimization}

\begin{figure}[t!]
\centering
\vspace{4pt}\includegraphics[width=0.95\columnwidth]{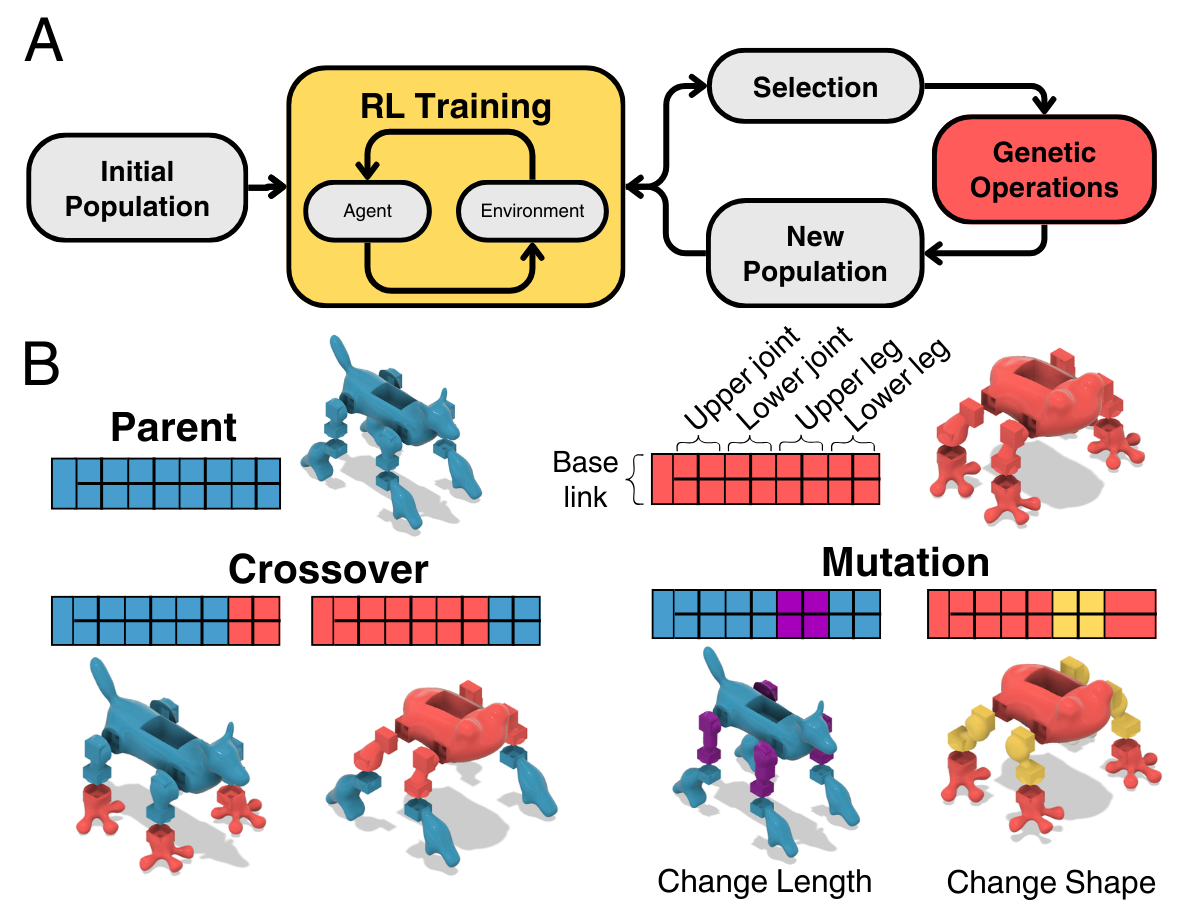}
\centering
\caption{\textbf{Morphology and Walking Policy Co-optimization.} (A) The inner loop implements reinforcement learning to optimize the robot control policy, and the outer loop optimizes the robot morphologies through genetic operations. (B) Our genetic representation and examples of crossover and mutation operation.}
\vspace{-30pt}
\label{fig: genetic operators}
\end{figure}

We propose an evolutionary algorithm (Fig.~\ref{fig: genetic operators}A) with a dual-loop architecture to optimize both robot morphology and control policy while simultaneously incorporating user preferences. The inner loop employs reinforcement learning to assess a robot's capacity for acquiring a walking policy. The outer loop utilizes genetic operators to evolve the robot morphology from a design repository.

An initial population of robots will be trained for locomotion to track changing velocities along $x-y-yaw$ directions. The robot observations include the previous actions, base linear and angular velocities, joint positions and velocities, the gravity vector projected onto the robot's coordinate system, and the target body velocity commands. The policy outputs the joint position offsets that will be converted to torques with a PD controller. The reward function consists of a baseline reward ($r_\textrm{baseline}$) and optional user-adjustable reward terms based on preference. $r_\textrm{baseline}$ minimizes velocity tracking error, maintains a plausible robot pose, and penalizes excessive joint torque, accelerations, frequent stepping behavior, and abrupt action changes. The user-specified components are the linear velocity tracking reward and joint power penalty. The per-step reward can be defined as:
\vspace{-2pt}
\begin{equation}
\label{eq: reward}
r =  \underbrace{\alpha_1 e^{-0.25{{\left\|\mathbf{v}^*_{xy}-\mathbf{v}_{xy}\right\|}^2}}}_{\text{linear velocity tracking}}
+ \underbrace{\alpha_2 \sum_{i=0}^{n} {\left\|\dot{q}_i\tau_i\right\|}}_{\text{joint power}} +r_\text{baseline}
\end{equation} 
where $\mathbf{v}^*_{xy}$ is the commanded base linear velocity in $xy$ direction, and $\mathbf{v}_{xy}$ is the actual base linear velocity,  $\dot{\mathbf{q}}$ is the joint velocities,  $\mathbf{\tau}$ is the joint torques,  $n$ is the number of joints and $\alpha_1$ and $\alpha_2$ are the weights of the respective reward terms. $r_\text{baseline}$ are listed in the supplementary material.

We select the top $100$ robots for each generation and create another $100$ new robots with genetic operators. 50 of the new robots are created through mutations, and the other 50 are created through crossovers. Crossover is achieved by duplicating a random robot in the current generation and choosing with a 50\% chance to swap a joint or a limb with another random robot in the current generation. For the mutations, the robot can undergo a change in limb length, limb shape, body shape, joint axis or remain the same with the possibility of 15\%, 15\%, 25\%, 40\%, and 5\%, respectively. A constraint of symmetry is imposed on the robot based on the observation of natural animal characteristics and the practicality of our implementation. To achieve this, the same change is applied to all legs or joints to unify the joint type and length or shape of the limb within the same body level. Although a larger design space can be achieved without this constraint, this ensures our robot has a stable starting pose at the beginning of the inner loop training and improves the training quality.

By defining the fitness score based on different evaluation metrics, our outer loop can prioritize different performances for robot selection and optimize both morphology and walking policy toward user preference. We design two criteria for users to prioritize: (1) Velocity tracking. We scale the velocity tracking reward by 20 and add it to the total reward. (2) Energy efficiency. We scale the energy penalty by 10 and add it to the total reward. 

\vspace{-5pt}
\mypara{Implementation Details} We trained our multi-directional walking policy using the Proximal Policy Optimization (PPO) algorithm \cite{schulman2017proximal}, following the implementation of parallel reinforcement learning described in \cite{rudin2022learning, rl-games2021}. Each robot was trained for $2.46 \times 10^7$ steps (a few minutes) before evaluation. We extended the open-source IsaacGymEnvs simulation environment \cite{makoviychuk2021isaac} for our parallel training with $4096$ environments per robot. Both our actor and critic networks employ three fully connected layers with dimensions of [512, 256, 128] and ELU activation functions. All training were performed on servers with NVIDIA RTX A6000, NVIDIA A100, and NVIDIA GeForce RTX 3090 GPUs. Comprehensive details on the specific training hyperparameters are in the supplementary material.

\subsection{Physical Assembly}

All robots were printed on the Creality CR-10 Smart Pro 3D printer due to its large printing bed. We use the Hiwonder HTD-45H High Voltage Serial Bus Servo to actuate our robots. We list detailed information on other electric components in supplementary material. The resulting robot assembly can be completed in \textbf{minutes} due to our modular design and careful considerations of electronic component placement and manufacturability. 

\section{Experiment}

In this section, we aim to evaluate Text2Robot from various aspects. First, we assess its ability to generate robot designs that align with diverse user-specified aesthetics. Second, we highlight the key advantages of integrating generative models into the automated robot design workflow. Third, we evaluate the performance of co-optimizing body morphology and control policy while prioritizing different performance metrics as specified by users. Finally, we demonstrate the real-world applicability of Text2Robot by fabricating and showcasing the physical robots.

\begin{figure}[t!]
\centering
\includegraphics[width=0.95\columnwidth]{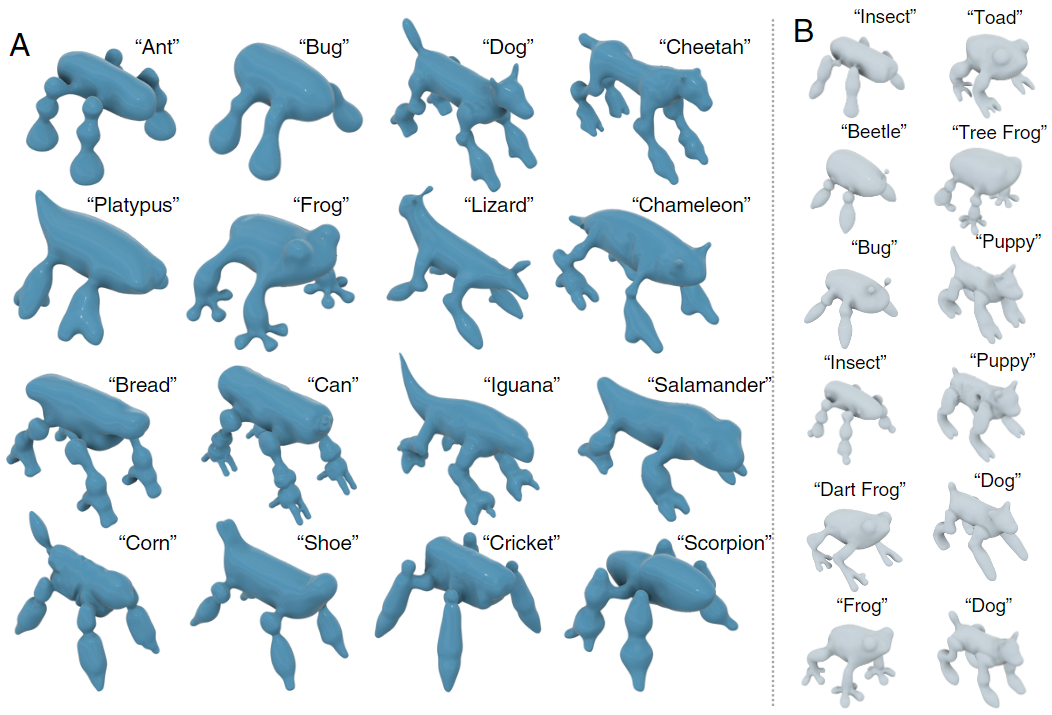}    
\centering
\caption{\textbf{Generated Meshes and Corresponding User Descriptions.} (A) Sixteen robot mesh models generated from our structured prompt with diverse user descriptions. (B) We used the same or similar descriptions to generate four other morphology variants for bug, frog, and dog robots. }
\label{fig: banks}
\vspace{-20pt}
\end{figure}

\subsection{Aesthetic Specification Matching}

To assess the effectiveness of Text2Robot in generating robot designs that align with diverse user-specified aesthetics, we evaluated the initial stages of the framework using a variety of input text prompts. This evaluation aimed to demonstrate the capabilities of our chosen generative model in the context of robot design. We tested sixteen diverse prompts, encompassing both animal-inspired descriptions (e.g., ``dog'', ``frog'') and more creative concepts (e.g., ``bread'', ``can'', ``shoe''). The qualitative results in Fig.~\ref{fig: banks}A demonstrate that the generated designs can capture the essence of the input descriptions while adhering to the fundamental structure of a quadrupedal robot. We further investigated achieving subtle variations in generated morphologies by tweaking prompts, producing unique robot designs for specific species like ``Bug'', ``Dog'', and ``Frog'', as shown in Fig.~\ref{fig: banks}B. For each of the above designs, we exported $30$ robots in total by our geometric processing stage with various leg lengths and joint orientations for subsequent experiments. Text2Robot can generate a robot mesh within $1\sim2~\si{mins}$ and convert it into a URDF model in $30~\si{secs}$.

\subsection{Advantages of Incorporating Text-to-3D Models}

\begin{figure}[t!]
\centering
\vspace{4pt}
\includegraphics[width=0.85\linewidth]{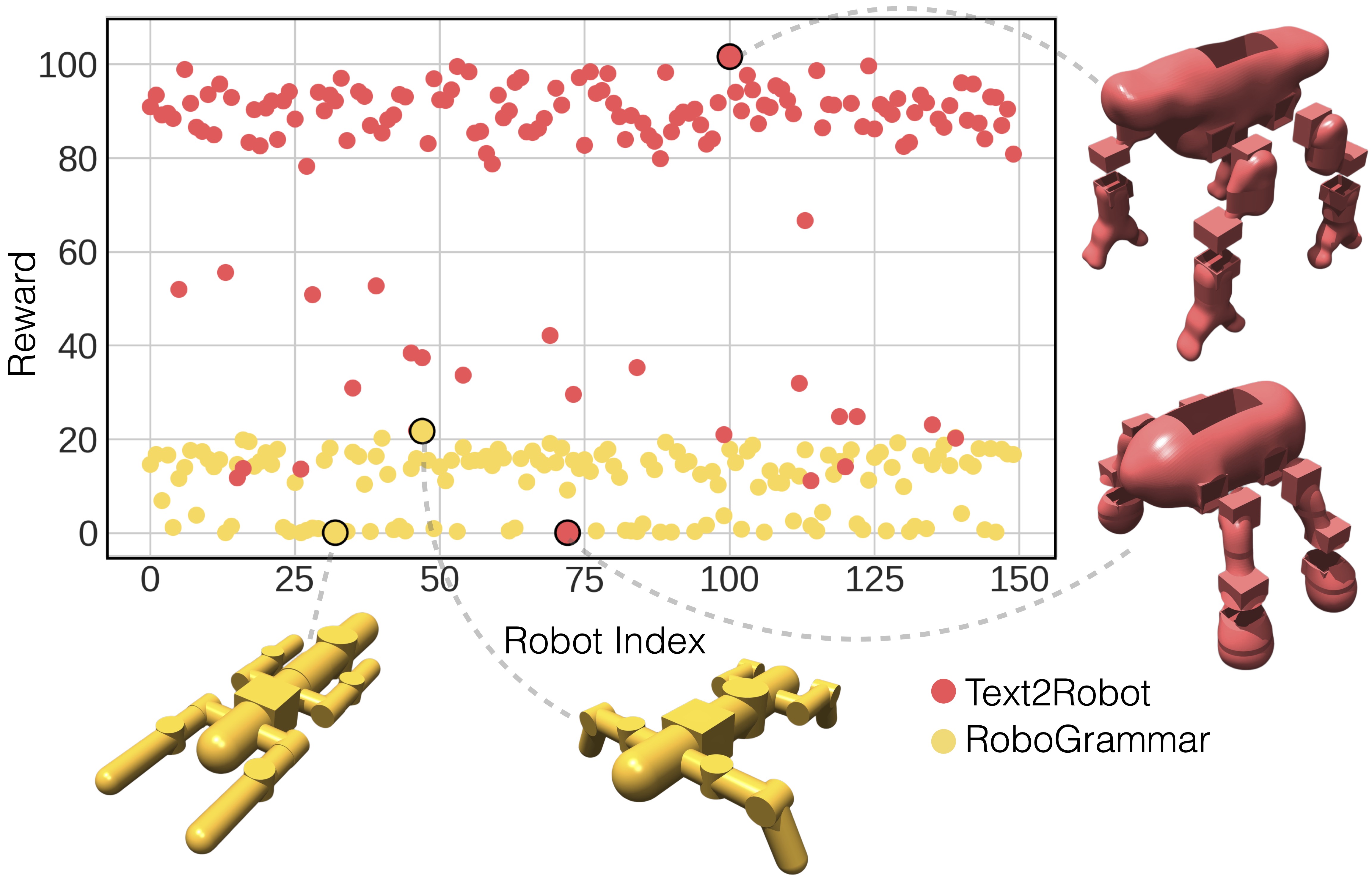}
\centering
\caption{Reward comparison of the Text2Robot and RoboGrammar robots and visualization of the best-performing and worst-performing robots.}
\vspace{-20pt}
\label{fig: baseline}
\end{figure}

We hypothesize that one key advantage of integrating text-to-3D models is to provide a much stronger initialization for evolutionary design. Therefore, we compared with RoboGrammar \cite{zhao2020robogrammar} which enables a wide range of symmetrical robot designs based on simple primitive geometries as a baseline. To ensure a fair comparison, we filtered robots generated with their recursive graph grammar to only include quadrupedal robots with two joints per leg and with similar body lengths. We created URDF files for their robot and assigned the weights to match the weights of our robots. We then trained 150 RoboGrammar robots and 150 of our generated designs for the same amount of steps for a walking policy. Our 150 robots are randomly selected from 600 robots augmented from the sixteen prompts (Fig.~\ref{fig: banks}A) and another four prompts in the bug species (Fig.~\ref{fig: banks}B). Fig.~\ref{fig: baseline} shows that our initialized robots outperformed the baseline initialized robots by a large margin.

We found most RoboGrammar robots suffer from unnatural placement and orientation of joints and links. Limbs on the same side are often too close together which compromises balance, and links protrude in the same direction as the joint axis, making the limb ineffective for locomotion. In contrast, Text2Robot leverages the knowledge of the physical world embedded within a text-to-3D generative model to produce quadrupedal robots with appropriate leg lengths, link and body proportions, and stable static initial postures.

\subsection{Morphology and Walking Policy Co-optimization}

\begin{figure}[t!]
\vspace{3pt}
\centering
\includegraphics[width=0.85\columnwidth]{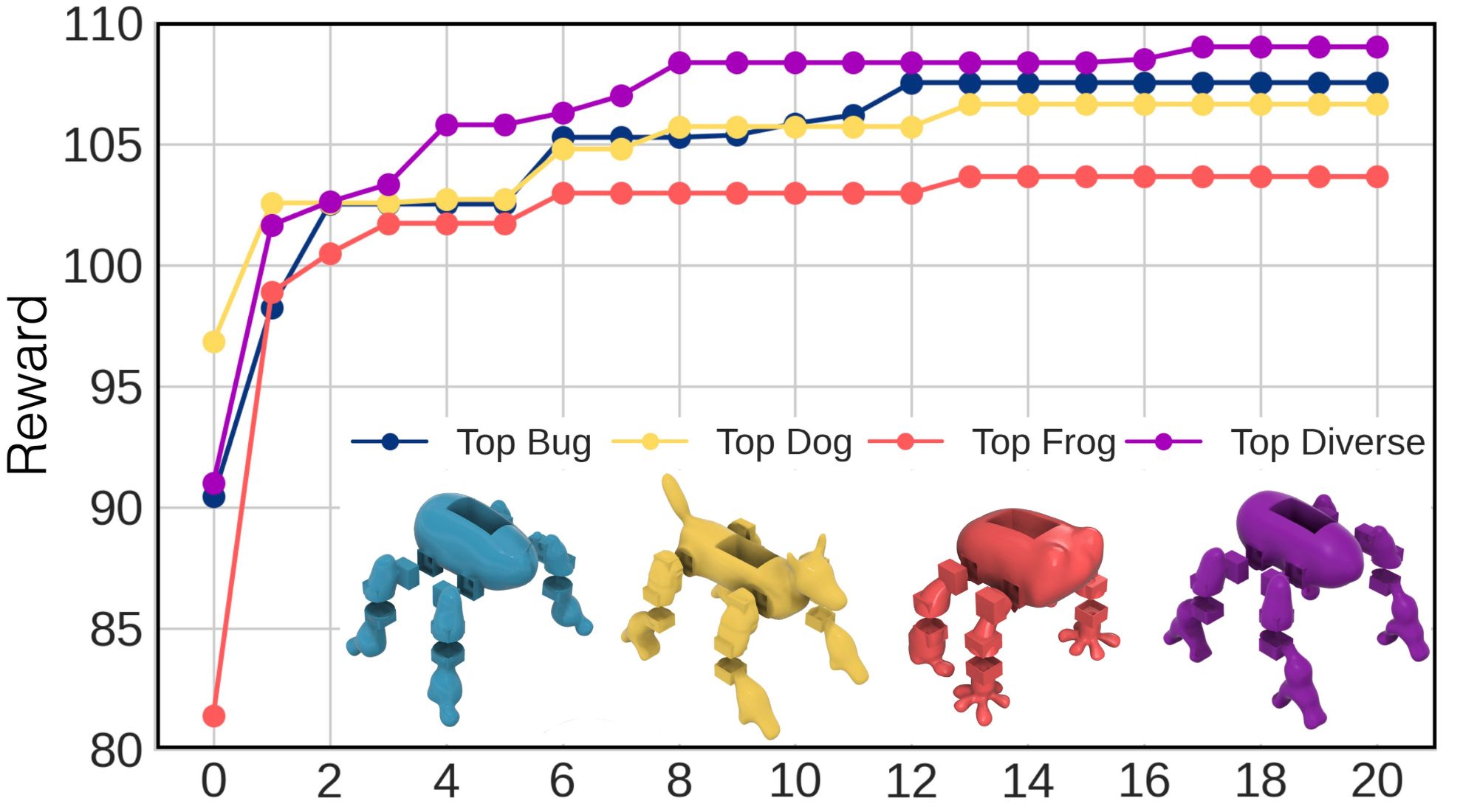}
\centering
\caption{\textbf{General Reward Optimization.} The reward of the best robot per generation and the morphology of the best robot in the last generation.}
\vspace{-20pt}
\label{fig: general reward}
\end{figure}

\addvspace{-4pt}

\mypara{Single Species Optimization} We first co-optimized morphology and control with similar morphologies to ``Bug'', ``Frog'' and ``Dog'' as shown in Fig~\ref{fig: banks}. Each species contains five variants of morphology from similar prompts. In each of the three trials, we used the 150 robots augmented from the five robots as the initial population in our EA, and all robots were optimized based on the general reward in Eq.~\ref{eq: reward}. Fig.~\ref{fig: general reward} shows the emergence of successful locomotion within only a few generations, indicating the effectiveness of our method in generating walking quadrupedal robots. We continued to evolve the robots for 20 generations, and the performance was further improved with evolved morphologies. 

\mypara{Increasing Diversity} We then investigated the effect of increasing diversity in the robot bank. We used the 600 robots augmented from the sixteen prompts and an additional four prompts in the bug species. The final selected robot achieved higher rewards compared to the robot optimized with similar morphologies (Fig.~\ref{fig: general reward}). Text2Robot enables higher-quality designs simply by expanding the diversity of text descriptions. This demonstrates the potential of our method to scale up and achieve better performance with more diverse and creative text prompts.

\subsection{Co-optimization with Preferences}

\begin{figure}[b!]
\vspace{-15pt}
\centering
\includegraphics[width=0.9\columnwidth]{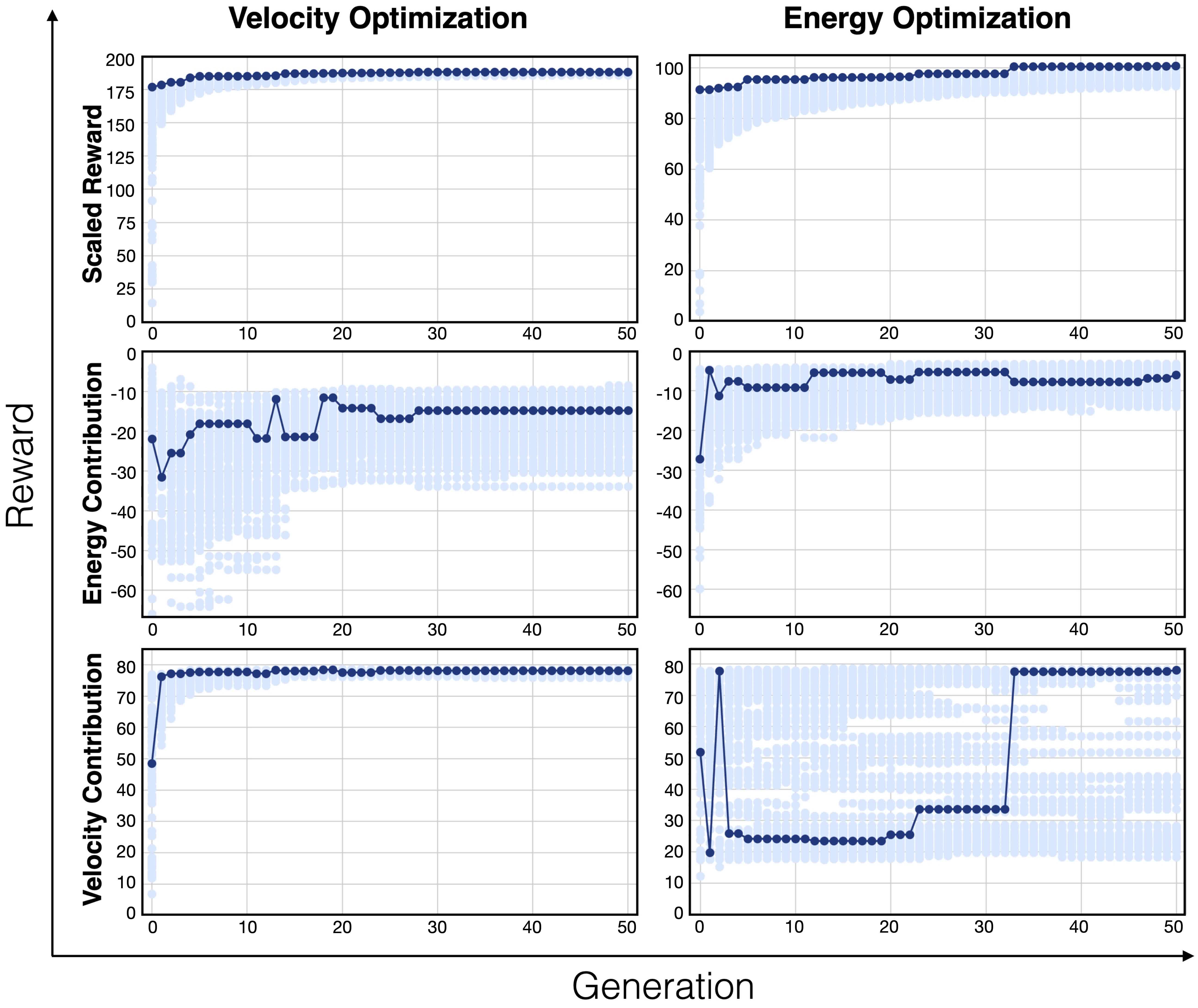}
\centering
\caption{\textbf{Optimization Based on Preference with Diverse Robot Species.} The scaled reward represents the fitness score evolutionary loop. It consists of the original reward received by the robots during inner loop training and the scaled contribution based on energy or velocity, depending on the user's preference. The best-performing robots are marked with dark blue, and the rest of the top 100 robots are marked with light blue in the figure.  }
\label{fig:performance_optimization}
\end{figure}

We evaluate Text2Robot to optimize robot designs according to user-supplied priorities: energy or velocity tracking. We used all robots with their augmented sets from Fig.~\ref{fig: banks}. We adjusted the calculation of the fitness score by scaling the reward with additional amplified energy or velocity contribution. Fig.~\ref{fig:performance_optimization} shows the results of diverse species based on each preference under 50 generations of evolution. The results of a single species are listed in the supplementary material.

In the optimization for robots from single species or diverse species, our method is able to optimize the robot performance per generation while considering the performance priorities. The results show a strong correlation between the top-performing robot and the targeted performance criteria, and the other performance criteria, which were not being optimized, appeared more random and sporadic. These results demonstrate that our robot is optimized to meet user preferences. We found the robots selected from velocity optimization generally have longer and wider bodies, while the robots selected from energy optimization have lower body weights. We also find that the performance of the diverse robot bank is better than that of single-species banks, indicating the scalability of our method. 

\subsection{Co-optimization for Rough Terrain}

We applied Text2Robot to rough terrain and observed that it effectively informed evolved robot morphologies with higher-performing foot shapes. We also aim to highlight the impact of input texts on informing details of the robot designs, such as foot shapes, beyond just the large body components as shown above. We used the robots from the diverse bank in Fig.~\ref{fig: banks} and applied a VHACD decomposition to increase the simulated realism of foot contact. Following prior curriculum design \cite{makoviychuk2021isaac}, robots are trained to progress through increasingly challenging terrains (smooth slope, rough slope, stairs, discrete, and stepping stones).

\begin{figure}[t!]
\vspace{5pt}
\centering
\includegraphics[width=0.95\columnwidth]{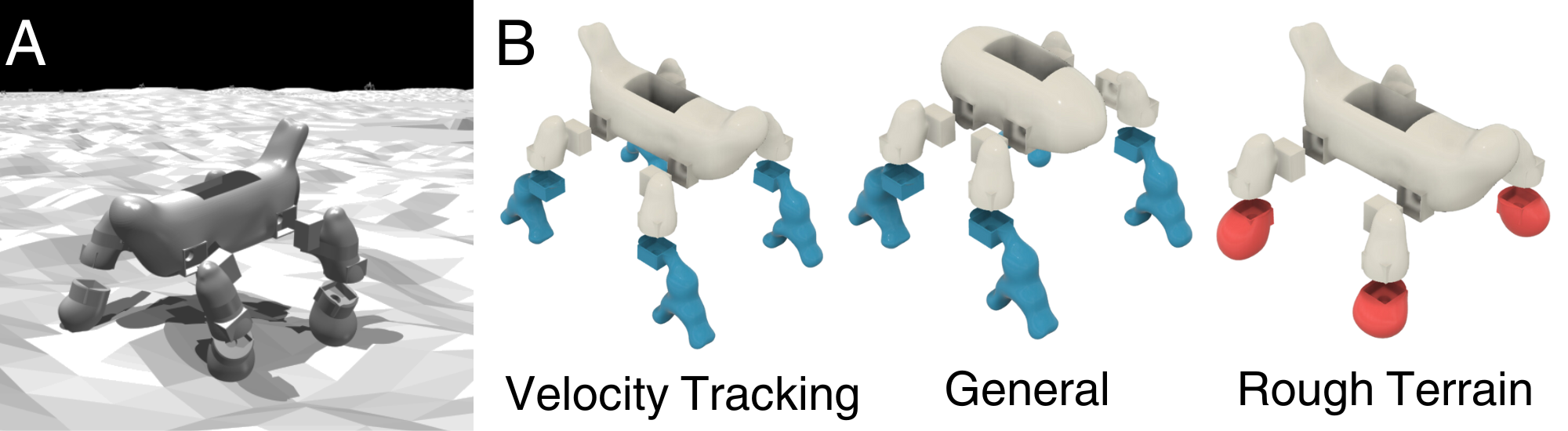}
\centering
\caption{\textbf{Rough Terrain Optimization.} (A) The selected robot traversing rough terrain. (B) Robots optimized for flat terrain (left and middle) evolve to have larger arcs for feet, while the robot evolved for rough terrain (right) has smaller, simple rounded feet.}
\vspace{-10pt}
\label{fig: rough terrain}
\end{figure}

Analysis of selected robot morphologies reveals a correlation between foot shape and terrain type, as illustrated in Fig.~\ref{fig: rough terrain}. Arched feet, favored in flat terrain trials, provide enhanced stability, speed, and efficient energy transfer due to their curvature. This advantage, however, hinges on predictable surface contact dynamics. We observed that the large foot dimensions of arched feet increase the risk of snagging on uneven terrain. Conversely, simple rounded feet, chosen for rough terrain, demonstrated superior adaptability to unpredictable surfaces, promoting stability and balance.
\vspace{-5pt}
\subsection{Physical Walking Robot}

\begin{figure}[t!]
\centering
\includegraphics[width=0.95\linewidth]{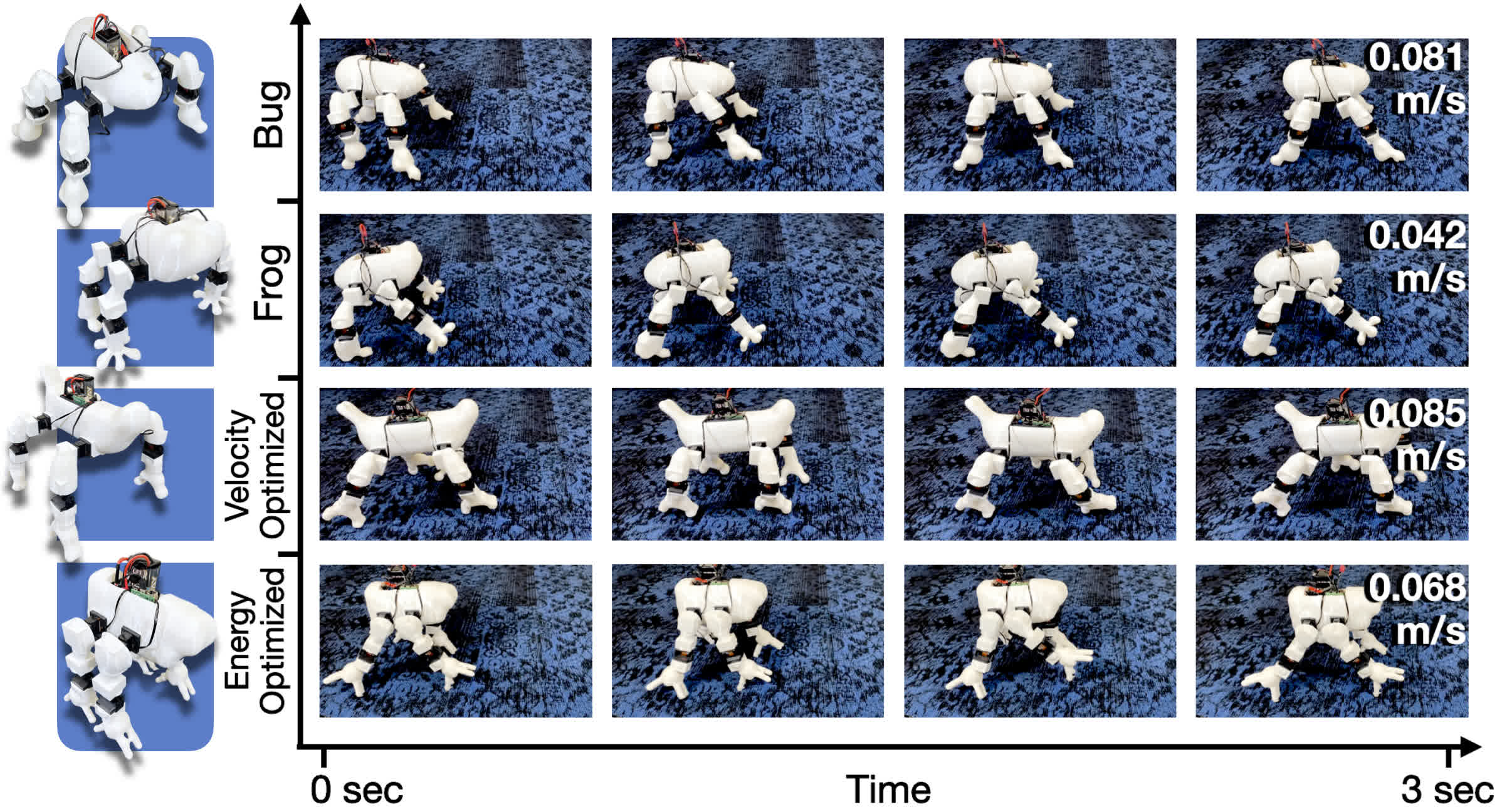}
\centering
\caption{\textbf{Real Robot Performance.} We play the best robot policy in simulation with a goal speed of $0.1~\si{m/s}$ in a straight direction, and the real robot executes the same position command in the real world.}
\vspace{-20pt}
\label{fig: sim2real}
\end{figure}

We selected the highest-performing ``Bug'' and ``Frog'' from the single-species optimization and the two robots from the diverse bank optimized for velocity tracking and energy efficiency to demonstrate our ability to fabricate the generated designs. Each robot required approximately a day to manufacture, but assembly (Fig.~\ref{fig: sim2real}) was completed within minutes. We show a primitive Sim2Real transfer by simply playing the trained locomotion in simulation and directly executing the joint positions on the real robot. As shown in Fig.~\ref{fig: sim2real}, our real robots successfully transfer the walking policy learned in simulation to the real world and achieve sufficient performance in locomotion and speed. This further validates the practicability and robustness of our design from our Text2Robot pipeline. More examples can be found in our supplementary videos.

\section{Conclusion, Limitation and Future Work}

We introduce Text2Robot, which generates a physical quadrupedal walking robot from text prompts to match user-specified aesthetic and performance preferences. Text2Robot leverages generative models for stronger initialization than traditional methods, while converting visual meshes to movable robots with considerations in electronics and real-world manufacturability. Both simulated and physical experiments show Text2Robot's ability to co-optimize morphology and control to produce physically functional machines.

\mypara{Limitations and Future Work} Text2Robot presents several opportunities for improvement in future research. While our current focus is on quadrupedal robots which are already challenging to design for humans, future work can extend the scope to robots with varying numbers of joints or other types of electromechanical machines. One possible solution is to combine the strong initialization of Text2Robot with existing robot design methods such as RoboGrammer. Additionally, the current framework still requires manual assembly. Integrating our method with automated assembly algorithms to construct physical robots would be a significant advancement. Furthermore, our text prompts remain fixed once optimization begins. Exploring a feedback mechanism to refine mesh generation from the text-to-3D model based on reward signals could offer greater flexibility.


\bibliographystyle{IEEEtran}
\bibliography{text2robot}

\clearpage

\section*{Supplementary Material}

\subsection*{A. Single Species Optimization}

\begin{figure}[h]
    \centering
    \includegraphics[width=0.9\linewidth]{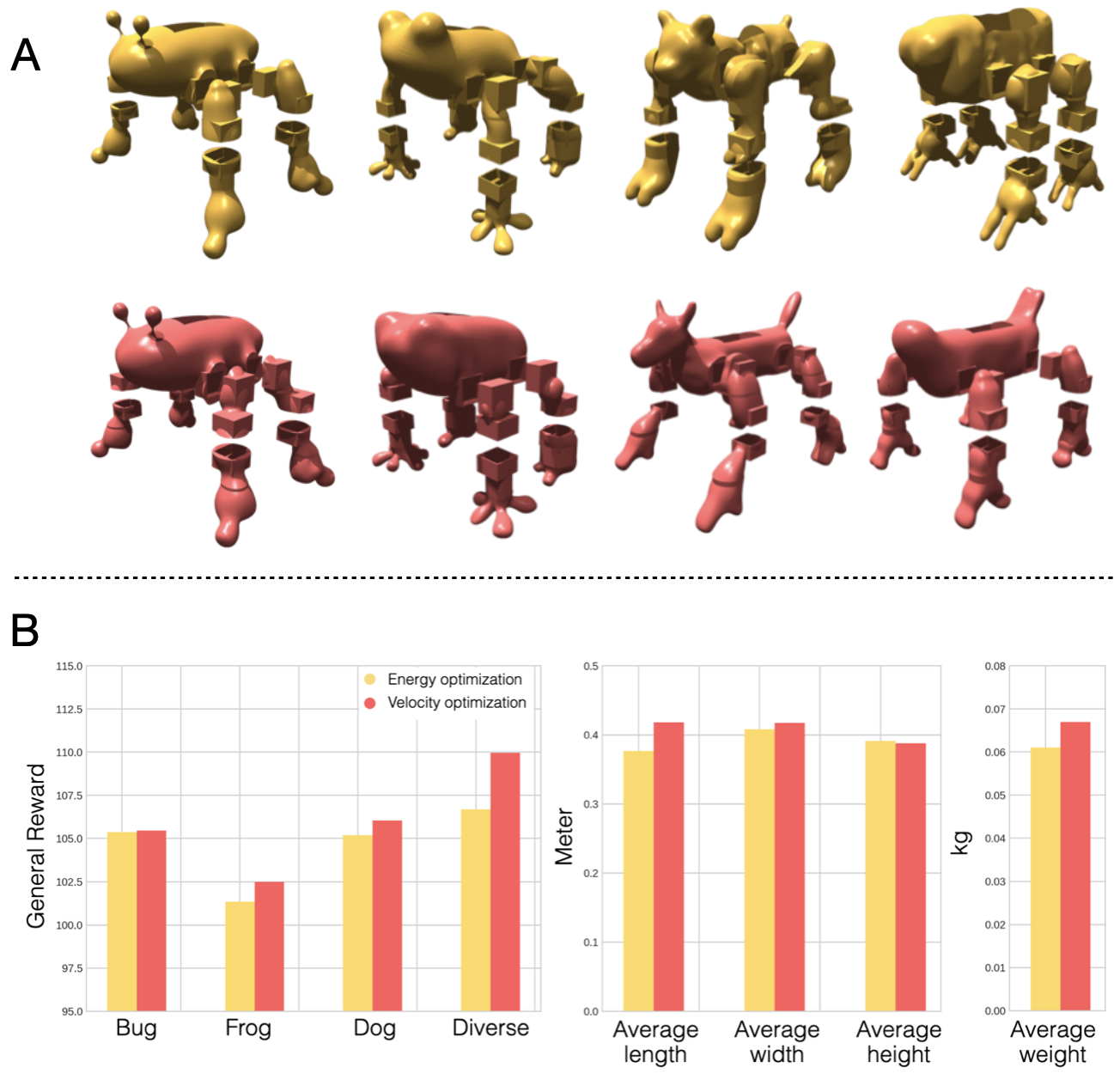}
    \caption{(\textbf{A})The morphology of the best robot in the last generation from the eight experiments with energy or velocity prioritize. Yellow color represents the robots that prioritize energy contribution and red color represents the robots that prioritize velocity contributions. (\textbf{B}) The unscaled reward of the eight best robots. The average length, width, height and weight of the eight robots. }
    \label{fig: total rewards}
\end{figure}

\begin{figure}[t!]
    \centering\includegraphics[width=1\linewidth]{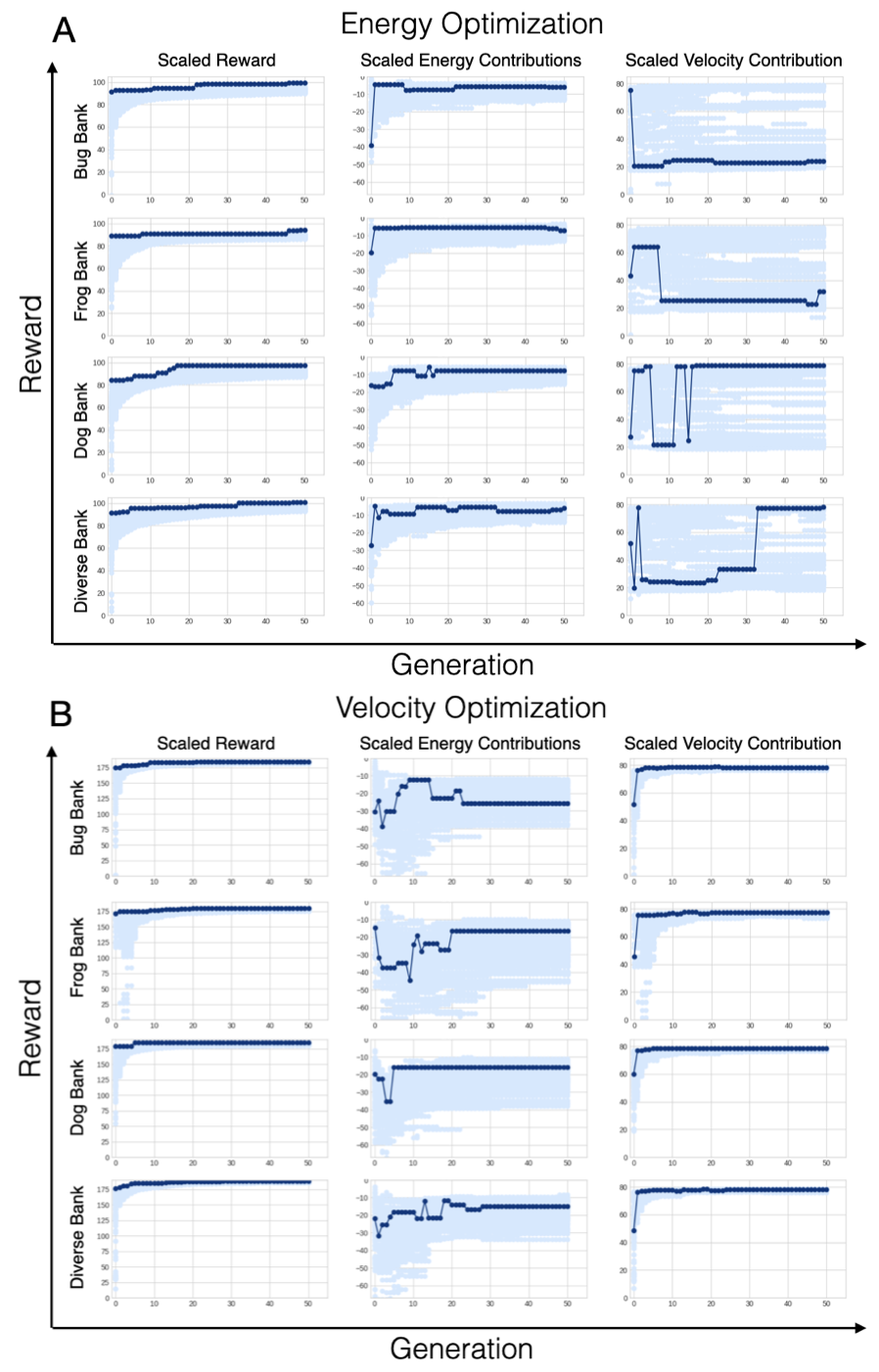}
    \caption{The scaled reward is used as the fitness metric in the EA loop. It consists of the original reward received by the robots during inner loop training summed with the requested priority: (\textbf{A}) Scaled energy contribution, (\textbf{B}) Scaled velocity contribution. The best-performing robots are marked with dark blue, and the rest of the top 100 robots are marked with light blue in the figure.}
    \label{fig: optimized breakdown}
\end{figure}

We show performance optimization results for single-species robot banks, 'Bug', 'Frog', and 'Dog,' and compare performance to that of the diverse bank. As in our diverse bank performance optimization experiment, we adjusted the fitness score calculation with an additional velocity reward or energy cost through 50 generations of evolution to demonstrate the effect of an input performance preference on final designs. As shown in Fig.~\ref{fig: optimized breakdown}, our results show a large correlation between the selected bot and the prioritized performance criteria. Fig.~\ref{fig: total rewards} shows the visualization of the selected robots, their average final rewards, and physical characteristics. 

\subsection*{B.\quad Reinforcement Learning Details}
We show the detailed training parameters of the inner reinforcement learning loop in Tab.~\ref{table:hyperparameters} , and the definition of symbols and baseline reward in Tab.~\ref{table:symbol_definition} and Tab.~\ref{table:baseline_reward}.

\subsection*{C.\quad Full Hardware Specifications}

We manufacture our robots using Creality CR-10 Smart Pro 3D printers. Robots are assembled using the 3D printed parts, as well as various electronic components which are easily inserted in the design. A comprehensive list of materials used in the construction of our robot is outlined in Table~\ref{table:components}.

\begin{table}[t!]
\centering
\small
\begin{tabular}{rc}
\toprule
Hyper-parameters & Values \\
\midrule
Dense network shape & [512, 256, 128] \\
Dense network activation & elu \\
Discount factor & 0.99 \\
GAE discount factor & 0.95 \\
PPO loss clip range & 0.2 \\
Entropy coefficient & 0.001 \\
Learning rate $\alpha$ & adaptive \\
Batch size & 98304 (4096x24)  \\
Mini-batch size & 16384 (4096x4) \\
Mini epochs & 5 \\
Critic loss coefficient & 2 \\
KL-divergence threshold & 0.008 \\
\bottomrule
\end{tabular}
\caption{PPO hyper parameters}
\label{table:hyperparameters}
\vspace{-10pt}
\end{table}

\clearpage
\begin{table}[!t]
\centering
\small
\begin{tabular}{rc}
\toprule
Base linear velocity & $\mathbf{v}$ \\
Base angular velocity  & $\omega$ \\
Commanded base linear velocity & $\mathbf{v}^*$ \\
Commanded base angular velocity & $\omega^*$ \\
Joint positions & $\mathbf{q}$ \\
Joint velocities & $\dot{\mathbf{q}}$ \\
Joint accelerations & $\ddot{\mathbf{q}}$ \\
Target joint positions & $\mathbf{q}^*$ \\
Joint torques & $\mathbf{\tau}$ \\
Number of joints & $n$ \\
Number of feet & $n_f$ \\
Feet air time & $t_{air}$ \\
Feet stance time & $t_{stance}$ \\
Base gravity & $\mathbf{g}_{b}$ \\
Environment time step & $dt$ \\
\bottomrule
\end{tabular}
\caption{Definition of symbols.}
\label{table:symbol_definition}
\end{table}

\begin{table}[!h]
\begin{tabular}{rcr}
\toprule
 baseline reward terms & definition & weight $[*dt]$ \\
\midrule
Linear velocity tracking & $e^{-0.25{{\left\|\mathbf{v}^*_{xy}-\mathbf{v}_{xy}\right\|}^2}}$ & $1 $ \\
Angular velocity tracking & $e^{-0.25{{\left\|\mathbf{\omega}^*_{z}-\mathbf{\omega}_{z}\right\|}^2}}$ & $0.5 $ \\
Linear velocity penalty & $v^2_{z}$ & $-4 $ \\
Angular velocity penalty & $\left\|\mathbf{\omega}_{xy}\right\|^2$ & $-0.05 $ \\
Joint acceleration penalty & $\left\|\ddot{\mathbf{q}}\right\|^2$ & $-5^{-7} $ \\
Joint torque penalty & $\left\|\mathbf{\tau}\right\|^2$ & $-2^{-5} $ \\
Action rate penalty & $\left\|\mathbf{\dot{a}}\right\|^2$ & $-5^{-5} $ \\
orientation & $\left\|\mathbf{g}_{b,xy}\right\|^2$  & -0.5 \\
Feet air time & $\sum_{k=0}^{n_f} (t_{air,k} - 0.5)$ & $0.1 $ \\
Feet stance time & $\sum_{k=0}^{n_f} (t_{stance,k} - 0.5)$ & $0.1 $ \\
\bottomrule
\end{tabular}
\caption{\textbf{Definition of baseline reward ($r_\text{baseline}$) terms.} The baseline reward contains base velocity and orientation tracking terms, action rate penalty, and joint torque penalty; the air time and stance time reward encourages longer air time and stand time to promote a more natural and fluid walking gait.}
\label{table:baseline_reward}
\end{table}

\begin{table}[h]
\centering
\captionsetup{justification=raggedright,singlelinecheck=false} 
\begin{tabularx}{\textwidth}{rXr}
\toprule
Component & Model & Quantity per robot \\
\midrule
Microcontroller & Raspberry Pi $4$ Model B & $1$ \\
Battery & Povway $5200\ \text{mA} \ \text{Lipo} \ 3S\  11.1V \ 50C$ & $1$ \\
Servo Motor & Hiwonder $HTD-45H$ High Voltage Serial Bus Servo $45\ \text{KG}$ & $8$ \\
DC to DC Power Converter & DROK $10A$ Synchronous Step-Down Voltage Regulator DC-DC $4-30V$ to $1.2-30V\ 12V$ & $1$ \\
PLA Filament & Ender PLA 3D Printer PLA Filament $1.75\ \text{mm} \ 1KG\ (2.2\ \text{lbs})\ \text{Spool PLA White}$ & $1$ \\
Motor Controller & Hiwonder TTL / USB Debugging Board & $1$ \\
\bottomrule
\end{tabularx}
\caption{\textbf{List of hardware component.}}
\label{table:components}
\end{table}

\end{document}